\documentclass[runningheads]{llncs}

\usepackage[T1]{fontenc}
\usepackage{graphicx}
\usepackage{placeins}
\usepackage{float}
\usepackage{booktabs}
\usepackage{array}
\usepackage{amsmath}
\usepackage{url}
\usepackage{adjustbox}
\usepackage{etoolbox}
\usepackage{tikz}
\usepackage{pgfplots}
\usetikzlibrary{arrows.meta,positioning,shapes.geometric,pgfplots.groupplots}
\pgfplotsset{compat=1.18}
\definecolor{raiteal}{HTML}{007C83}
\definecolor{raiorange}{HTML}{D94F04}

\DeclareGraphicsExtensions{.pdf,.png,.jpg,.jpeg}
\newcolumntype{P}[1]{>{\raggedright\arraybackslash}p{#1}}
\newcommand{\noact}{a^{\mathrm{noop}}}
\providecommand{\Description}[1]{}
\AtBeginDocument{%
  }

\begin{document}
\raggedbottom

\title{Runtime Action Interference for AI Control of \textit{AlphaStar} in \textit{StarCraft~II}} 
\titlerunning{Runtime Action Interference for AI Control}

\author{Jaymari Chua\inst{1,2} \and
Chen Wang\inst{2} \and
Liming Zhu\inst{1,2} \and
Lina Yao\inst{1}}

\authorrunning{Chua et al.}

\institute{University of New South Wales, Sydney, Australia\\
\and
CSIRO, Australia\\}

\maketitle

\begin{abstract}
A trained reinforcement learning policy does not determine the complete behavior that users encounter: deployment code still schedules, admits, suppresses, or replaces its proposed actions. We contribute \emph{runtime action interference} (RAI), an AI control mechanism that preserves policy parameters while regulating action pacing and filtering configured action patterns after inference. RAI releases a proposed action only when its cooldown condition is satisfied and its content detector does not flag the action; otherwise, it dispatches a no-op. The detector covers specified toxic behaviors, including worker-unit harassment, while the cooldown controls action rate. We implement RAI in a replication of AlphaStar actor.py and make the implementation and reproducibility materials available through an open source code repository. We deployed RAI in a \textit{StarCraft~II} human participant study that compared two presentations of the same opponent with high capability and rate limited actions; we withheld its capability claim in one presentation and disclosed it in the other. On response scales from 1 to 5, we observed pooled fairness, trust, and toxicity means of 3.90, 3.50, and 2.00 under claim withholding, compared with 2.62, 4.31, and 2.85 under disclosure. Disclosure corresponded with lower perceived fairness and higher perceived toxicity across every expertise group, whereas trust increased among novices and experts but decreased among intermediate participants. Our human evaluation therefore shows that perceptions of an opponent controlled through RAI can vary substantially with the capability information presented to users, even when the configured control remains constant. We conclude that human-computer evaluations must separate control within the execution stack from capability disclosure and assess fairness, trust, and toxicity as distinct dimensions of human experience.
\end{abstract}

\section{Introduction}

Learned policies propose actions, but interactive systems determine which proposals become observable behavior. Schedulers, rate limits, queues, and interface constraints operate in the actor stack after policy inference and before action dispatch to the environment. This part of the stack determines which proposed actions reach the game and when they execute. Rank and win rate characterize policy capability, while players encounter the action sequence produced after these deployment controls have been applied.

We introduce \emph{runtime action interference} (RAI) to make this boundary explicit. RAI places a deterministic admission rule between policy inference and dispatch: a proposal reaches the environment only when it satisfies the safety condition and passes the action-content detector; otherwise, it becomes a no-op. The mechanism leaves the policy parameters fixed and can therefore be integrated without retraining the policy. In our implementation, the detector identifies configured toxic action patterns, including worker-unit harassment, while the cooldown regulates action timing. RAI therefore combines timeframe control with targeted content filtering rather than treating every inferred action as eligible for execution.

\begin{figure}[H]
  \centering
  \includegraphics[width=\textwidth,keepaspectratio]{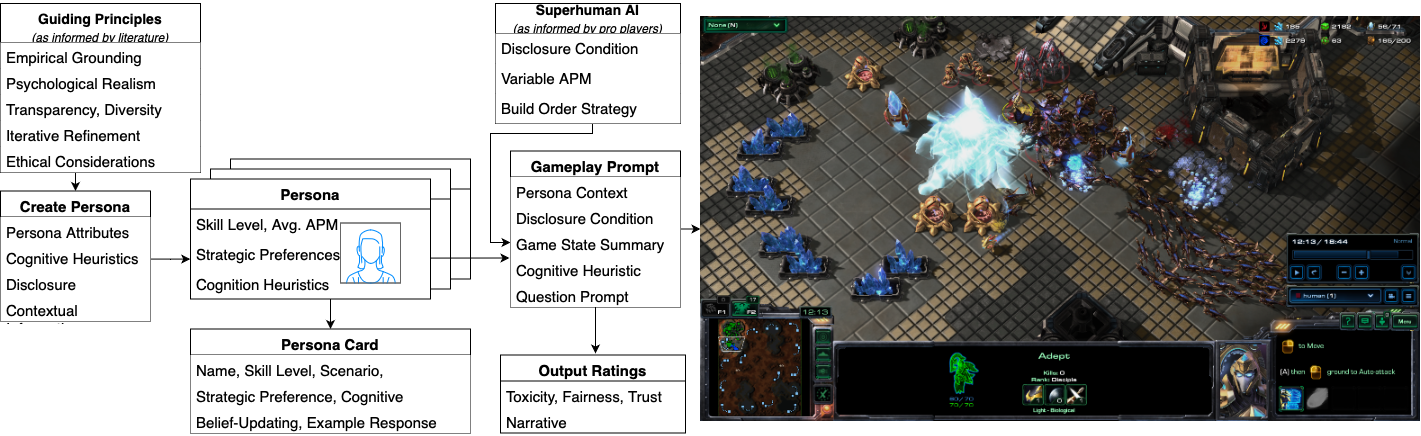}
  \caption{Study context linking personas of human participants with the \textit{StarCraft~II} gameplay, and post-match responses.}
  \Description{The figure places Persona Card and prompting materials beside StarCraft II gameplay and the fairness, trust, toxicity, and narrative response dimensions.}
  \label{fig:study-overview}
\end{figure}

We used RAI in a \textit{StarCraft~II} human-participant study. We retained 32 post-match responses across three reported conditions. The focal 23 responses concern the same documented high-capability, rate-controlled opponent: we withheld its capability claim for 10 responses and disclosed it for 13. We retain the others as baseline. We then evaluated both what RAI allows the agent to execute and how players perceive the resulting opponent. We report the following: capability information can shape those perceptions even when the controller remains unchanged. Trust is not identical to reliance and depends on the person, task, and interaction history \cite{lee2004trust,hoff2015trust,swinger2025teammait}. Fairness captures whether players regard the competition as level, while toxicity captures whether they experience the encounter as hostile or aversive.

We ask two research questions. \emph{RQ1: How can an existing reinforcement learning policy be regulated after inference without modifying its parameters, and what records make that runtime interference auditable?} \emph{RQ2: What descriptive differences in fairness, trust, toxicity, and narrative interpretation appear when users encounter the same documented RAI configuration with its capability claim withheld or disclosed?}

We contribute (1) a model of action admission after policy inference; (2) a code implementation of such audit contract that connects policy proposals, controller decisions, and released actions in \textit{StarCraft II}; and (3) a human evaluation of fairness, trust, toxicity, and narrative interpretation around an opponent documented as using the shared controller.

\section{Related Work}

\subsection{Control After Policy Inference}

We position RAI at a different stage from methods that constrain policy learning. Constrained policy optimization incorporates restrictions into the learning objective \cite{achiam2017cpo}, while runtime shielding blocks or replaces actions that violate a specification \cite{alshiekh2018shielding}. Other recent methods modify a policy during or after training: TARL updates selected parameters during testing to reduce action uncertainty under distribution shift, whereas GGPO learns explicit safety costs during policy optimization \cite{xu2025tarl,chua2026ggpo}. RAI instead operates after inference and before the actor dispatches an action to the environment. It leaves the trained policy fixed, checks each proposal for timeframe eligibility and configured toxic action patterns, releases proposals that pass both checks, and substitutes a no-op otherwise. We formalize this control point, specify the records required to audit it, and distinguish the configured controller from the capability information presented to participants.

RAI constitutes \emph{self-repair applied after inference as a system}. Action proposals issued by the policy is not presumed to be executable: the controller may withhold it or defer its release, and it may instead dispatch a substitute action whose effects conform to the deployment specification. Automated program repair likewise diagnoses defective program behaviour and synthesizes a validated modification to the program that produced it \cite{bouzenia2025repairagent}. RAI preserves this corrective logic while relocating its point of application from source code or policy parameters to the stream of actions produced during execution. AgentSpec demonstrates how such separation can be implemented through rules that intercept proposed agent actions before their effects reach the environment \cite{wang2026agentspec}. As such, these approaches recast the central design question: which behavioural obligations should be entrusted to policy learning, and which should remain externally specified so that they can be inspected and revised during deployment?

\subsection{Execution Constraints in Competitive AI}

The evaluation of competitive agents already depends on execution constraints. AlphaStar limited action rates and camera access, while chess research has optimized agents to resemble human play rather than treating unconstrained strength as the sole objective \cite{vinyals2019grandmaster,mcilroy2020aligning}. Benchmark fairness similarly depends on whether people and artificial agents receive comparable information and action resources \cite{canaan2019leveling}. Recent research in human computer interaction extends this concern to competition among participants with different abilities, the disclosure of assistance, and the framing of AI participation in competitive settings \cite{trindade2025mixed,sako2025assistance,pan2026fair}. These studies motivate an account of both policy capability and the behavior that the deployment stack permits users to encounter.

A related tradition in systems design places verification before execution, thereby treating an inferred action as a candidate for review rather than as an inevitable consequence of the model's output. AgentSpec realizes this separation through rules that are evaluated while the agent is running \cite{wang2026agentspec}. Reinforced Agent adopts a related arrangement, but delegates the review of a proposed tool call to an independently configurable component that can intervene before execution \cite{ta2026reinforced}. In both cases, the behaviour exposed to the environment emerges from the interaction between a model proposal and the mechanism that governs admission; if admission fails, the prescribed fallback determines the resulting action. RAI carries this architectural separation into competitive play through a deterministic verifier of restricted scope, e.g. framerates and definitions of toxic game actions, while its event record makes each intervention observable at the level of an individual action.

\subsection{Capability Framing and Competitive Perceptions}

We do not assume that transparency produces uniform benefits. Explanations can increase acceptance without improving complementary performance, and recent interventions can reduce excessive reliance without consistently improving appropriate reliance \cite{bansal2021explanations,bo2025rely}. Studies of disclosures about AI use and visualizations of human and AI contributions further show that transparency can reduce perceived legitimacy or alter judgments of contribution, depending on the context and presentation format \cite{schilke2025dilemma,kusters2026disclosure}. Expectations of trust and collaboration vary with expertise and role, while uncertainty displays can reshape trust and reliance as people gain experience with a system \cite{swinger2025teammait,chen2025missing}. In competition, disclosed assistance can become part of the fairness judgment itself \cite{sako2025assistance}. Judgments of toxicity likewise depend on social context and can range from severe harm to routine annoyance \cite{beres2021toxicity,laato2024toxicity}. We therefore preserve the distinctions among fairness, trust, and toxicity rather than combining them into a single measure of acceptance.

\section{Method}
\label{sec:methods}

\subsection{Runtime Action Interference}

We implemented RAI in RAI-Alphastar actor.py, where it operates after policy inference and before dispatch to the environment. Let $h_t$ denote the policy history, and let the fixed policy propose $a_t^{\pi}\sim\pi_\theta(\cdot\mid h_t)$. The controller maintains cooldown state $c_t$, evaluates timeframe eligibility through $E_\phi(c_t,t)$, and applies content detector $T_\psi(a_t^{\pi},h_t)$, where $T_\psi=1$ denotes a configured toxic action pattern:
\begin{equation}
  \widetilde{a}_t=
  \begin{cases}
    a_t^{\pi}, & E_\phi(c_t,t)=1 \ \land\ T_\psi(a_t^{\pi},h_t)=0,\\
    \noact, & \text{otherwise},
  \end{cases}
  \qquad c_{t+1}=U_\phi(c_t,t,\widetilde{a}_t).
  \label{eq:rate-gate}
\end{equation}
We keep the policy parameters $\theta$ fixed throughout deployment. The controller parameters $\phi$ govern timeframe admission and state evolution, while detector configuration $\psi$ identifies prohibited action patterns such as worker-unit harassment. A proposal executes only when both predicates permit release. RAI consequently changes the executed action process without altering the policy that generated the proposal. Should either predicate fail, the proposal is vetoed and superseded by $\text{no\_act}$ (a benign null-operation). 

To govern these proposals particularly, and preclude infinite intervention loops, a subordinate controller maintains a stateful cooldown registry, $c_t$. Action admissibility is arbitrated via two orthogonal predicates:
\begin{enumerate}
    \item \textbf{Timeframe Eligibility ($E_\phi(c_t,t)$):} Dictates execution permissibility based on configuration $\phi$ and the prevailing cooldown state of the game.
    \item \textbf{Content Detector ($T_\psi(a_t^{\pi},h_t)$):} Scrutinizes the proposed action against historical context via configuration $\psi$, flagging prohibited behavioral paradigms.
\end{enumerate}

Hence, RAI intercedes solely at the terminal output layer; while policy parameters $\theta$ remain strictly immutable.

\begin{figure}[H]
  \centering
  \begin{tikzpicture}[
      node distance=2mm,
      every node/.style={font=\footnotesize,align=center},
      block/.style={draw,rounded corners,minimum height=9mm,inner sep=1mm},
      flow/.style={-{Latex[length=2mm]},semithick},
      audit/.style={draw,dashed,rounded corners,inner sep=2mm,text width=.84\textwidth}
    ]
    \node[block,text width=18mm] (policy) {Fixed policy\\$\pi_\theta(\cdot\mid h_t)$};
    \node[block,right=of policy,text width=16mm] (proposal) {Proposal\\$a_t^\pi$};
    \node[block,right=of proposal,text width=22mm] (eligibility) {RAI checks\\$E_\phi(c_t,t)$ and $T_\psi(a_t^\pi,h_t)$};
    \node[block,right=of eligibility,text width=22mm] (execution) {Released action\\$a_t^\pi$ or $\noact$};
    \node[block,right=of execution,text width=19mm] (environment) {Environment};
    \draw[flow] (policy) -- (proposal);
    \draw[flow] (proposal) -- (eligibility);
    \draw[flow] (eligibility) -- (execution);
    \draw[flow] (execution) -- (environment);
    \node[audit,below=7mm of eligibility] (record) {Audit record: timestamp, policy proposal, cooldown decision, detector decision and reason, released action, and controller state.};
    \draw[dashed] (proposal.south) -- (record.north west);
    \draw[dashed] (eligibility.south) -- (record.north);
    \draw[dashed] (execution.south) -- (record.north east);
  \end{tikzpicture}
  \caption{RAI releases a proposal only when it satisfies the cooldown condition and passes the configured content detector; otherwise, RAI substitutes a no-op. The audit record captures both decisions.}
  \Description{The fixed policy produces a proposal. Runtime Action Interference checks cooldown eligibility and configured toxic-action patterns before releasing the proposal or substituting a no-op.}
  \label{fig:rate-gate}
\end{figure}
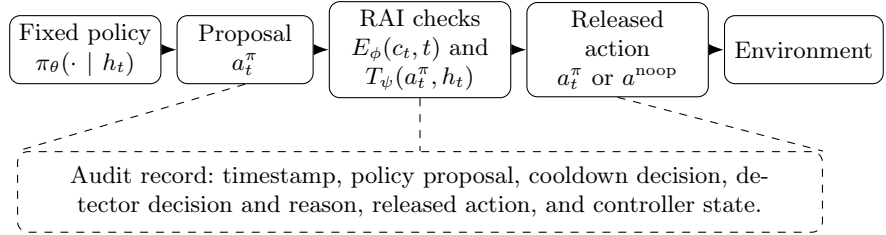
\FloatBarrier

In the actor path, we combine target APM cooldown control, no-op substitution, and content detection. The detector flags configured toxic patterns, including worker-unit harassment, while the cooldown limits the rate at which otherwise eligible proposals may execute. Equation~\ref{eq:rate-gate} and Figure~\ref{fig:rate-gate} specify the resulting decision rule and the event fields required to distinguish temporal interventions from content interventions.

RAI-AlphaStar also contains a renderer utility that clears queued interface commands. Neither the user nor the agent can instantiate that renderer; queue sanitation in the renderer is therefore implemented separately from RAI in \texttt{actor.py}.

\subsection{Human Study Design}

We conducted a human participant study in which players encountered an AI opponent in \textit{StarCraft~II}, completed post-match ratings, and provided a narrative response. The study included three conditions and 32 responses. Our focal design compares two presentations of the same configured opponent with high capability and RAI enabled. We withheld the capability claim in one presentation and disclosed it in the other, thereby varying the information available to participants while holding the opponent and controller configuration constant. The focal comparison contains 10 responses under claim withholding and 13 under disclosure.

We stratified participants by prior expertise. We classified novices as new to \textit{StarCraft~II}; intermediates ranked within the top 50\% regionally in comparable real-time strategy games but did not meet the expert threshold; and experts ranked within the top 10\%. Under claim withholding, the novice, intermediate, and expert counts were 2, 4, and 4. Under disclosure, the corresponding counts were 5, 3, and 5. We conducted the study under institutional ethics approval.

Table~\ref{tab:conditions} separates the controller configuration from the capability information presented to participants.

\begin{table}[H]
  \caption{Focal human study design. Both presentations used the same configured opponent and RAI controller; we varied whether the capability claim was withheld or disclosed. Counts are novice/intermediate/expert (N/I/E).}
  \label{tab:conditions}
  \centering
  \small
  \begin{tabular}{@{}P{0.21\textwidth}cP{0.31\textwidth}P{0.22\textwidth}@{}}
    \toprule
    Presentation & $n$ (N/I/E) & Configured opponent & Capability information \\
    \midrule
    Baseline & 9 (4/2/3) & Baseline; RAI inactive & No capability claim reported \\
    Claim withheld & 10 (2/4/4) & High capability; RAI active in the actor & Claim withheld \\
    Claim disclosed & 13 (5/3/5) & High capability; RAI active in the actor & Claim disclosed \\
    \bottomrule
  \end{tabular}
\end{table}
\FloatBarrier

\subsection{Measures}

We measured fairness, trust, and toxicity on scales from 1 (lowest) to 5 (highest). \emph{Fairness} concerned whether the opponent's behavior was consistent with community expectations given its disclosed or undisclosed capability. \emph{Trust} concerned willingness to play another match or rely on the opponent's gameplay insights. \emph{Toxicity} concerned an aversive interaction ranging from overt aggression to subtle condescension. We treated the three measures as distinct outcomes because confidence in an opponent does not imply that the encounter was regarded as fair or non-toxic.

\subsection{Descriptive Analysis}

For every combination of outcome, presentation, and expertise, we analyze the cell mean, descriptive standard deviation, variance, and sample size reported in Appendix~\ref{app:human-aggregates}. For each outcome, we calculate (i) condition means weighted by expertise cell size, (ii) the disclosed minus withheld difference within each expertise group, and (iii) a contrast standardized to the combined focal distribution of 7 novices, 7 intermediates, and 9 experts:
\begin{equation}
  \Delta_{\mathrm{std}}=
  \frac{7\Delta_N+7\Delta_I+9\Delta_E}{23}.
  \label{eq:composition}
\end{equation}
As a sensitivity analysis for group composition, we also assign equal weight to the three expertise tiers, $\Delta_{\mathrm{eq}}=(\Delta_N+\Delta_I+\Delta_E)/3$. We interpret all contrasts descriptively because the focal cells contain between two and five responses.

\subsection{Code and Data Availability}

We provide the aggregate human data used in our analysis in Appendix~\ref{app:human-aggregates}, together with the cell means, descriptive standard deviations, variances, and sample sizes needed to reproduce every reported contrast. We also provide the RAI implementation, its formal decision rule, and the audit schema shown in Figure~\ref{fig:rate-gate}. The nine responses from the third condition remain available as aggregate data, but we exclude them from the focal comparison.

The open source dataset produced through this research contains participant-level ratings, the assignment and exclusion ledger, and the verbatim questionnaire. We also include the exact disclosure script and the RAI controller settings required to reproduce the presentation and execution conditions. These materials support participant-level inspection and reproducibility; however, because the expertise cells remain small, the present analysis is descriptive and does not make causal claims from the condition contrasts.

For the runtime layer, the code repository includes launch manifests, replays, proposal logs, detector decisions, executed-action logs, and match-level telemetry from the participant sessions. These records permit calculation of realized intervention frequencies and verification of the action streams presented in each condition.

\section{Results}
\label{sec:results}

\subsection{Ratings and Participant Accounts}

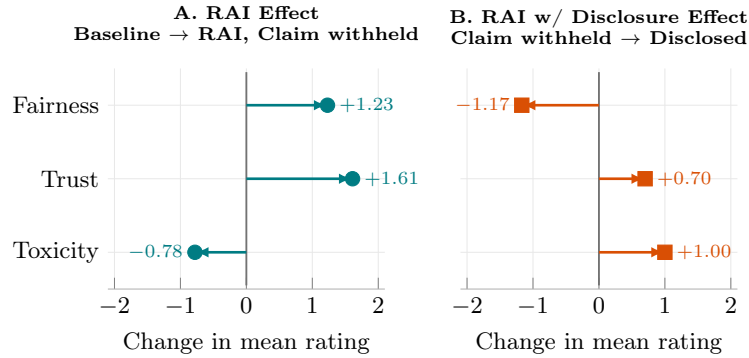
\begin{figure}[t]
  \centering
  \begin{tikzpicture}
    \begin{groupplot}[
      group style={group size=2 by 1,horizontal sep=10mm,y descriptions at=edge left},
      width=.43\linewidth,
      height=4.5cm,
      xmin=-2.1,xmax=2.1,
      ymin=.5,ymax=3.5,
      xtick={-2,-1,0,1,2},
      ytick={1,2,3},
      yticklabels={Toxicity,Trust,Fairness},
      tick label style={font=\rmfamily\small},
      label style={font=\rmfamily\small},
      axis y line*=left,
      y axis line style={draw=none},
      ytick style={draw=none},
      axis x line*=bottom,
      xmajorgrids,
      ymajorgrids,
      grid style={gray!18,line width=.25pt},
      axis line style={gray!60},
      title style={font=\rmfamily\scriptsize\bfseries,align=center},
      xlabel={Change in mean rating},
      clip=false
    ]
      \nextgroupplot[title={A. RAI Effect\\Baseline $\rightarrow$ RAI, Claim withheld}]
        \draw[black!55,line width=.7pt] (axis cs:0,.55) -- (axis cs:0,3.45);
        \addplot[-{Latex[length=1.8mm,width=1.4mm]},raiteal,line width=1pt,forget plot] coordinates {(0,3) (1.23,3)};
        \addplot[-{Latex[length=1.8mm,width=1.4mm]},raiteal,line width=1pt,forget plot] coordinates {(0,2) (1.61,2)};
        \addplot[-{Latex[length=1.8mm,width=1.4mm]},raiteal,line width=1pt,forget plot] coordinates {(0,1) (-0.78,1)};
        \addplot[only marks,mark=*,mark size=2.7pt,raiteal] coordinates {(1.23,3) (1.61,2) (-0.78,1)};
        \node[anchor=west,font=\rmfamily\scriptsize\bfseries,raiteal] at (axis cs:1.28,3) {$+1.23$};
        \node[anchor=west,font=\rmfamily\scriptsize\bfseries,raiteal] at (axis cs:1.66,2) {$+1.61$};
        \node[anchor=east,font=\rmfamily\scriptsize\bfseries,raiteal] at (axis cs:-.83,1) {$-0.78$};

      \nextgroupplot[title={B. RAI w/ Disclosure Effect\\Claim withheld $\rightarrow$ Disclosed},yticklabels=\empty]
        \draw[black!55,line width=.7pt] (axis cs:0,.55) -- (axis cs:0,3.45);
        \addplot[-{Latex[length=1.8mm,width=1.4mm]},raiorange,line width=1pt,forget plot] coordinates {(0,3) (-1.17,3)};
        \addplot[-{Latex[length=1.8mm,width=1.4mm]},raiorange,line width=1pt,forget plot] coordinates {(0,2) (.70,2)};
        \addplot[-{Latex[length=1.8mm,width=1.4mm]},raiorange,line width=1pt,forget plot] coordinates {(0,1) (1.00,1)};
        \addplot[only marks,mark=square*,mark size=2.7pt,raiorange] coordinates {(-1.17,3) (.70,2) (1.00,1)};
        \node[anchor=east,font=\rmfamily\scriptsize\bfseries,raiorange] at (axis cs:-1.22,3) {$-1.17$};
        \node[anchor=west,font=\rmfamily\scriptsize\bfseries,raiorange] at (axis cs:.75,2) {$+0.70$};
        \node[anchor=west,font=\rmfamily\scriptsize\bfseries,raiorange] at (axis cs:1.05,1) {$+1.00$};
    \end{groupplot}
  \end{tikzpicture}
  \caption{Two contrasts in participant ratings. Panel A reports pooled differences between the separately reported baseline and the RAI condition with the capability claim withheld. These differences are descriptive and do not identify a causal effect of RAI because the baseline implementation was not recovered. Panel B reports composition-standardized disclosed-minus-withheld differences under the same configured RAI controller. Positive values denote higher ratings on the corresponding outcome.}
  \Description{Panel A shows higher fairness and trust and lower toxicity for the RAI claim-withheld condition relative to the reported baseline. Panel B shows lower fairness and higher trust and toxicity under capability disclosure with the RAI controller held constant.}
  \label{fig:human-results}
\end{figure}

We recorded user participant ratings as each expertise by presentation cell contains between two and five responses. Although the released dataset contains participant-level records, the small cells do not support stable hypothesis tests or precise interval estimates. We therefore report means, descriptive standard deviations, and arithmetic contrasts without claims of statistical significance. Table~\ref{tab:focal-ratings} presents every focal cell on the original response scale, while Figure~\ref{fig:human-results} shows the direction and magnitude of the disclosed minus withheld contrasts. These contrasts compare separate participant groups and do not represent within-participant change.

\begin{table}[H]
  \caption{Exploratory ratings under the shared RAI configuration. Cells report $M$ (descriptive SD) [$n$] on scales from 1 to 5; $\Delta$ is disclosed minus withheld. Pooled rows report sample-size-weighted means and omit SDs.}
  \label{tab:focal-ratings}
  \centering
  \small
  \setlength{\tabcolsep}{4pt}
  \begin{tabular}{@{}llccc@{}}
    \toprule
    Outcome & Expertise & Claim withheld & Claim disclosed & $\Delta$ \\
    \midrule
    Fairness & Novice & 3.50 (0.50) [2] & 2.20 (1.17) [5] & $-1.30$ \\
             & Intermediate & 4.25 (1.30) [4] & 2.67 (0.47) [3] & $-1.58$ \\
             & Expert & 3.75 (0.43) [4] & 3.00 (0.89) [5] & $-0.75$ \\
             & \textit{Pooled} & 3.90 [10] & 2.62 [13] & $-1.28$ \\
    \midrule
    Trust & Novice & 3.50 (1.50) [2] & 4.80 (0.40) [5] & $+1.30$ \\
          & Intermediate & 3.75 (0.83) [4] & 3.00 (1.41) [3] & $-0.75$ \\
          & Expert & 3.25 (1.48) [4] & 4.60 (0.49) [5] & $+1.35$ \\
          & \textit{Pooled} & 3.50 [10] & 4.31 [13] & $+0.81$ \\
    \midrule
    Toxicity & Novice & 1.00 (0.00) [2] & 3.00 (1.67) [5] & $+2.00$ \\
             & Intermediate & 2.50 (0.50) [4] & 3.00 (0.82) [3] & $+0.50$ \\
             & Expert & 2.00 (0.71) [4] & 2.60 (1.36) [5] & $+0.60$ \\
             & \textit{Pooled} & 2.00 [10] & 2.85 [13] & $+0.85$ \\
    \bottomrule
  \end{tabular}
\end{table}

\FloatBarrier

\paragraph{Fairness.}
Fairness was lower under disclosure in every expertise stratum. The disclosed minus withheld contrasts were $-1.30$ for novices, $-1.58$ for intermediates, and $-0.75$ for experts; the pooled means were 3.90 under claim withholding and 2.62 under disclosure. Standardization to the combined expertise distribution yielded $\Delta_{\mathrm{std}}=-1.17$, while equal weighting of the three expertise tiers yielded $-1.21$. A participant under claim withholding described the opponent as ``fair enough as there doesn't seem to be cheating,'' despite also remarking on its speed. This account provides an example of how perceived rule compliance entered a fairness judgment, even if it does not establish how frequently participants used that reasoning.

\paragraph{Trust.}
Trust did not follow a common direction across expertise groups. It was higher under disclosure among novices ($\Delta=+1.30$) and experts ($\Delta=+1.35$), but lower among intermediates ($\Delta=-0.75$). The pooled means were 3.50 under claim withholding and 4.31 under disclosure; composition standardization yielded $\Delta_{\mathrm{std}}=+0.70$, and equal weighting yielded $+0.63$. The positive pooled contrast therefore coexists with an expertise-specific reversal. Under disclosure, one participant explained that ``the AI gameplay was strong'' and consequently adopted a two-base strategy, while another referred directly to ``the disclosure of its past performance.'' Under claim withholding, a participant described the opponent as ``adaptive enough.'' These accounts show that participants attended to disclosed capability and observed play when discussing the opponent, but they do not explain the contrasting intermediate-group mean.

\paragraph{Toxicity.}
Toxicity was higher under disclosure in every expertise stratum, with contrasts of $+2.00$ for novices, $+0.50$ for intermediates, and $+0.60$ for experts. The pooled means were 2.00 under claim withholding and 2.85 under disclosure. Standardization yielded $\Delta_{\mathrm{std}}=+1.00$, and equal weighting yielded $+1.03$. The shared direction across weighting schemes indicates that the aggregate pattern is not produced solely by the different expertise composition of the two presentations.

Therefore, the standardized contrasts preserve lower fairness and higher toxicity under disclosure, while the trust contrast summarizes responses that move in opposing directions across expertise groups.

\section{Discussion}
\label{sec:discussion}

In RAI, parameter preservation must not be conflated with preservation of either strategy or capability. Once the controller suppresses a proposal or replaces it with a no-op, the deployed system implements a different closed-loop policy from the policy represented by the original weights alone. The substituted action changes the subsequent game state and therefore the history on which later proposals depend. It may also induce repeated proposals for an action that remains blocked, interrupt a strategic sequence, or reduce task performance. RAI is consequently parameter preserving, but it is not preserving of capability outcomes in interactive systems.

The controller and its configured prohibitions were part of the experimental deployment. In particular, the content detector was configured to suppress specified patterns such as worker-unit harassment, while the permissible action rates governed the release of otherwise admissible actions. This establishes that participants encountered an opponent operating through the RAI execution path; it does not, without proposal and execution logs, establish how often the controller intervened or how strongly those interventions altered play. Evidence of realised efficacy would require the rate at which proposals were delayed, suppressed, or replaced, together with task-performance measures under the controlled and uncontrolled execution paths. Replay analysis was used to determine whether a prohibited behaviour was genuinely reduced or instead displaced into a different sequence of actions with comparable effects.

As a limitation to this study, the human study should be interpreted as exploratory evidence about capability framing. Both focal presentations used the same RAI controller, so their contrast cannot show that RAI improved fairness, increased trust, or reduced perceived toxicity. Instead, the results describe how participants evaluated the same controlled opponent when its capability claim was withheld or disclosed. This is a consequential distinction: the study evaluates interpretation of a deployed control configuration, not the causal effect of introducing that control. We also explicitly scope the human evaluation as a probe into capability framing under a fixed runtime configuration, rather than a sufficiently powered psychological experiment. As both focal presentations used the identical RAI controller, the observed variances in fairness and toxicity still do highlight a critical vulnerability in interactive AI evaluation: user perception is highly volatile and heavily anchored by capability disclosure, even when deterministic guardrails are active, reinforcing the necessity of the RAI architecture.

A deployment claim about a controlled policy should describe the behavior users can encounter, not only the policy checkpoint or nominal controller setting. The event record in Figure~\ref{fig:rate-gate} connects each proposal to its cooldown decision, detector decision, released action, timestamp, and controller state, permitting calculation of rate interventions, content interventions, and executed action timing. Linking those records to replays and participant outcomes would turn configured release rules into an auditable account of realized control.

\section{Conclusion}

RAI provides a concrete control point after policy inference: cooldown admission, configured toxic-action detection, and no-op substitution regulate which proposals reach the environment without retraining the policy. We formalize that mechanism and the event record needed to audit it, then report participant perceptions around a \textit{StarCraft~II} opponent documented as using the shared RAI mechanism. Under claim withholding, pooled fairness, trust, and toxicity means were 3.90, 3.50, and 2.00; under disclosure, they were 2.62, 4.31, and 2.85. Fairness was lower and toxicity higher in every expertise stratum, while trust changed direction across strata. Since RAI was shared, these contrasts evaluate capability framing around the controlled opponent even if not entirely evident of RAI's causal effect. The broader contribution is an evaluation suite for human-computer interaction: trained policy capability, runtime-controlled behavior, and user interpretation are distinct objects that must be specified, measured, and connected through observed evidence in interactive software systems.

\bibliographystyle{splncs04}
\bibliography{paper-references}

\clearpage
\appendix
\begingroup
\BeforeBeginEnvironment{tabular}{\begin{adjustbox}{max width=\textwidth}}
\AfterEndEnvironment{tabular}{\end{adjustbox}}
\section{Aggregate Results}
\label{app:human-aggregates}

We report our complete condition-by-expertise aggregates in Table~\ref{tab:human-results-appendix}.

\begin{table*}[!ht]
  \caption{We report aggregate summaries from our $N=32$ human experiment. Expertise cells show $M$ (SD) [$n$] on five-point scales, and overall cells show sample-size-weighted $M$ [$n$]. We omit overall SDs that do not reconcile with the expertise-cell summaries.}
  \label{tab:human-results-appendix}
  \centering
  \small
  \begin{tabular}{llcccc}
    \toprule
    Outcome & Opponent configuration & Novice & Intermediate & Expert & Overall \\
    \midrule
    Fairness & Human--No Disclosure (reported baseline) & 3.25 (1.09) [4] & 2.00 (0.00) [2] & 2.33 (1.25) [3] & 2.67 [9] \\
             & Rate-limited high capability, claim withheld & 3.50 (0.50) [2] & 4.25 (1.30) [4] & 3.75 (0.43) [4] & 3.90 [10] \\
             & Rate-limited high capability, claim disclosed & 2.20 (1.17) [5] & 2.67 (0.47) [3] & 3.00 (0.89) [5] & 2.62 [13] \\
    \midrule
    Trust & Human--No Disclosure (reported baseline) & 2.25 (1.30) [4] & 2.00 (1.00) [2] & 1.33 (0.47) [3] & 1.89 [9] \\
          & Rate-limited high capability, claim withheld & 3.50 (1.50) [2] & 3.75 (0.83) [4] & 3.25 (1.48) [4] & 3.50 [10] \\
          & Rate-limited high capability, claim disclosed & 4.80 (0.40) [5] & 3.00 (1.41) [3] & 4.60 (0.49) [5] & 4.31 [13] \\
    \midrule
    Toxicity & Human--No Disclosure (reported baseline) & 2.25 (1.64) [4] & 2.50 (1.50) [2] & 3.67 (0.94) [3] & 2.78 [9] \\
             & Rate-limited high capability, claim withheld & 1.00 (0.00) [2] & 2.50 (0.50) [4] & 2.00 (0.71) [4] & 2.00 [10] \\
             & Rate-limited high capability, claim disclosed & 3.00 (1.67) [5] & 3.00 (0.82) [3] & 2.60 (1.36) [5] & 2.85 [13] \\
    \bottomrule
  \end{tabular}
\end{table*}

We retained means, reported descriptive standard deviations, variances, and cell sizes for every combination of presentation and expertise. We used the same RAI configuration in both focal presentations, including its temporal release and configured content-detection rules, and withheld the capability claim in the nondisclosure presentation. We did not retain individual ratings, the assignment and exclusion ledger, the verbatim questionnaire, the numeric study-specific cap, or match-level telemetry. We therefore use the table for descriptive orderings and arithmetic contrasts rather than participant-level diagnostic checks.

In the main text, we weight the disclosed-minus-withheld differences for novices, intermediates, and experts by our combined focal counts of 7, 7, and 9, respectively. Thus, for outcome $y$, we calculate $\Delta_{\mathrm{std}}=(7\Delta_{y,N}+7\Delta_{y,I}+9\Delta_{y,E})/23$. From the displayed cell means, we obtain $-1.17$ for fairness, $+0.70$ for trust, and $+1.00$ for toxicity, subject to rounding of the retained means.

\subsection{Condition Mapping}

We used the same documented rate-limited opponent and RAI configuration in both focal presentations and varied whether we withheld or provided the capability claim; we did not manipulate runtime interference. We did not retain the numeric target and realized APM, RAI event logs, launch configuration, checkpoint and replay identifiers, exact disclosure script, participant-level assignments, or baseline implementation. We treat the optional renderer queue reset as distinct from RAI in the actor and do not claim that it formed part of the participant study.

\FloatBarrier
\FloatBarrier
\section{Personas}
\label{app:synthetic-audit}

We report the quantitative summaries we retained from our synthetic \textit{StarCraft~II} pilot. We do not treat these outputs as human-subject data, combine them with our human aggregates, or use them to estimate effects in a player population. We did not retain raw generations, run identifiers, group sizes, complete prompts, random seeds, sampling settings, or generation logs. We therefore report the available summaries without claiming generation-level reproduction or validation.

We generated these outputs with \texttt{gemini-exp-\allowbreak1206} using a structured prompt schema containing persona identity, expertise, strategic preference, disclosure condition, game-state context, a named cognitive heuristic, and requested one-to-five ratings with a justification. Because we did not retain the complete run-level prompts, we cannot establish which settings produced each aggregate.

We labeled the synthetic comparison ``No Disclosure,'' although our materials separately name a constrained baseline and a high-capability nondisclosure configuration. We also collapsed novice and intermediate cards in the reported tests. Because we did not retain a run-level mapping, we cannot reconstruct the synthetic design unambiguously.

\begin{table*}[!ht]
\centering
\caption{We report our synthetic \textit{StarCraft~II} comparisons as model-generated outputs rather than participant observations.}
\label{tab:synthetic-sc2}
\small
\begin{tabular}{llccccc}
\toprule
Outcome & Reported expertise & No disclosure $M$ (SD) & Disclosure $M$ (SD) & $U$ & Reported $p$ & Reported $d$ \\
\midrule
Toxicity & Novice--Intermediate & 4.25 (0.55) & 2.45 (0.51) & 395.5 & $<.001$ & 3.39 \\
Toxicity & Expert & 5.00 (0.00) & 2.70 (0.57) & 400.0 & $<.001$ & 5.69 \\
Fairness & Novice--Intermediate & 1.85 (0.67) & 3.65 (0.49) & 10.5 & $<.001$ & $-3.07$ \\
Fairness & Expert & 1.55 (---) & 3.45 (---) & --- & $<.0001$ & --- \\
Trust & Novice--Intermediate & 2.45 (0.69) & 4.50 (0.51) & 0.0 & $<.001$ & $-3.38$ \\
Trust & Expert & 1.70 (0.73) & 4.35 (0.59) & 1.5 & $<.001$ & $-3.99$ \\
\bottomrule
\end{tabular}
\end{table*}

Because we did not retain a raw-log mapping or run identifiers, we do not treat the synthetic text fragments as a qualitative corpus.

\subsection{Implications for Use}

We analyze ancilliary study on the LLMs to predict persona materials separately from participant evidence. Recent research warns that substituting language model responses for participants can erase consent, agency, and contextual depth, while a review of generative persona research identifies substantial variation in how personas are constructed and evaluated \cite{kapania2025simulacrum,amin2026personas}. However, they can be useful in human computer interaction study design and therefore report Persona cards and their retained summaries as ancillary materials rather than treating them as evidence from participants.

We find that plausible synthetic outputs cannot expand our human evidence base. Our generated means and rationales depend on predefined expertise categories, named heuristics, prompts, and model settings, while our retained summaries leave unresolved contradictions in condition labels, grouping, and result direction. We therefore use these outputs prospectively to identify competing predictions, test output formats, and expose missing measures before participant recruitment. We do not use them to supply inferential power, participant variability, or quotations absent from our human data.
\endgroup
\FloatBarrier
\begingroup
\makeatletter
\setlength{\@fptop}{0pt}
\setlength{\@fpsep}{10pt plus 1fil}
\setlength{\@fpbot}{0pt plus 1fil}
\makeatother
\section{Data and Materials Availability}

We provide our retained synthetic summaries, Persona Card transcriptions, deidentified excerpts, and prompt documentation in the appendices. We did not retain participant-level ratings and do not release them publicly.

\subsection{Separating Control and Framing Effects}

A factorial follow-up should vary RAI and capability disclosure independently. An RAI-on/RAI-off comparison, combined with exact disclosure scripts, launch manifests, replays, event logs, and participant-level outcomes, would estimate whether runtime interference changes user perceptions, whether framing changes interpretations of a fixed action process, and whether the two factors interact. Such a design would also permit direct tests of whether different cooldown settings and content-detection rules preserve strategic challenge while reducing specified toxic actions.

\section{Ethics and Societal Impact Statement}

We obtained ethics approval and debriefed participants, but we did not retain the approval identifier or underlying review record in the current materials. We deidentified the excerpts and screened them to reduce re-identification risk. We keep synthetic outputs separate from participant evidence and interpret them as products of our Persona Cards, prompts, and model behavior rather than as substitutes for players.

\section{StarCraft II Persona Card Transcriptions}
\label{app:persona-data}

We transcribe our six detailed \textit{StarCraft~II} Persona Cards in this appendix and exclude cards from our separate chat-assistant scenario because they fall outside this competitive-game study. We did not retain the row-level synthetic interactions, so these transcriptions do not constitute a complete release of our Persona Cards dataset.

Capability statements should likewise refer to the deployed opponent, including material release constraints, rather than policy rank alone. In our data, a capability claim coincided with outcome-specific evaluations even though the documented technical configuration was shared. This result does not identify a causal mechanism, but it shows why policy capability, release control, and participant-facing information require separate specification.

\FloatBarrier

\begin{table}[tbp]
\centering
\caption{Persona Card SC2\_Novice\_1 (novice, defensive macro profile).}
\label{tab:app-sc2-novice-1}
\small
\begin{tabular}{p{0.21\linewidth}p{0.71\linewidth}}
\toprule
Attribute & Description \\
\midrule
Name & SC2\_Novice\_1 \\
Scenario & \textit{StarCraft~II} \\
Skill level & Novice \\
APM range & 35--50 (average: 45) \\
Description & A player new to RTS games, particularly StarCraft II. Prefers a defensive ``turtle'' playstyle, focusing on building a strong economy and large army before attacking. \\
Strategic preference & Turtle, Macro Focus \\
Cognitive heuristics & \textbf{Availability Heuristic:} Recent losses make the persona more likely to play defensively. The persona overreacts to recent events, particularly losses. \newline \textbf{Anchoring:} The persona tends to stick to its initial build order, even if scouting information suggests it's being countered. They are slow to adapt to new information or change their initial strategies. \newline \textbf{Confirmation Bias:} The persona interprets in-game events in a way that confirms their belief that a strong economy is the key to victory. They may downplay the importance of early aggression. \\
Belief updating & Slow to update beliefs about the opponent's strategy. Overweighs early game experiences. \\
Example response & ``I lost the last game because I didn't have enough units. This time, I'm going to focus even more on building up my economy before I attack. That's always the best way to win.'' \\
\bottomrule
\end{tabular}
\end{table}

\begin{table}[tbp]
\centering
\caption{Persona Card SC2\_Novice\_2 (novice, aggressive rush profile).}
\label{tab:app-sc2-novice-2}
\small
\begin{tabular}{p{0.21\linewidth}p{0.71\linewidth}}
\toprule
Attribute & Description \\
\midrule
Name & SC2\_Novice\_2 \\
Scenario & \textit{StarCraft~II} \\
Skill level & Novice \\
APM range & 30--50 (average: 35) \\
Description & A player new to RTS games, particularly StarCraft II. Prefers an aggressive playstyle, often attempting early rushes. \\
Strategic preference & Aggressive, Rush Focus \\
Cognitive heuristics & \textbf{Availability Heuristic:} Recent successes with rushes lead to more rushes. The persona overreacts to recent events, particularly successes. \newline \textbf{Anchoring:} The persona tends to attempt early rushes even when scouting suggests it is a bad idea. They are slow to adapt to new information or change their initial strategies. \newline \textbf{Overconfidence:} The persona overestimates their chances of a rush succeeding, often leading to early losses. \\
Belief updating & Slow to update beliefs, especially regarding the effectiveness of rushes. Overweighs early game experiences. \\
Example response & ``I won the last game with a 6-pool! It's the best strategy. I'm going to do it again and win even faster this time!'' \\
\bottomrule
\end{tabular}
\end{table}

\begin{table}[tbp]
\centering
\caption{Persona Card SC2\_Intermediate\_1 (intermediate, balanced profile).}
\label{tab:app-sc2-intermediate-1}
\small
\begin{tabular}{p{0.21\linewidth}p{0.71\linewidth}}
\toprule
Attribute & Description \\
\midrule
Name & SC2\_Intermediate\_1 \\
Scenario & \textit{StarCraft~II} \\
Skill level & Intermediate \\
APM range & 80--120 (average: 90) \\
Description & A player with some experience in RTS games, comfortable with basic strategies but still developing a deeper understanding of StarCraft II. Prefers a balanced playstyle, adapting to the opponent. \\
Strategic preference & Balanced, All-Rounder \\
Cognitive heuristics & \textbf{Adaptive Heuristic:} The persona attempts to switch between macro and aggressive play based on scouting information. They adjust their strategy based on observed opponent actions, with a delay. \newline \textbf{Representativeness:} The persona assumes the opponent will follow standard build orders and strategies. They make assumptions about the opponent's strategy based on limited information. \newline \textbf{Mental Accounting:} The persona values units produced earlier in the game more highly than those produced later, even if they are identical. \\
Belief updating & Moderately responsive to new information. Adapts to opponent's strategy but may be slow to recognize novel tactics. \\
Example response & ``Okay, they're going for an early Barracks. I'll build a few extra units and try to scout what they're doing. I should be able to hold if I don't overcommit.'' \\
\bottomrule
\end{tabular}
\end{table}

\begin{table}[tbp]
\centering
\caption{Persona Card SC2\_Intermediate\_2 (intermediate, harassment profile).}
\label{tab:app-sc2-intermediate-2}
\small
\begin{tabular}{p{0.21\linewidth}p{0.71\linewidth}}
\toprule
Attribute & Description \\
\midrule
Name & SC2\_Intermediate\_2 \\
Scenario & \textit{StarCraft~II} \\
Skill level & Intermediate \\
APM range & 80--120 (average: 110) \\
Description & A player with some experience in RTS games, comfortable with basic strategies but still developing a deeper understanding of StarCraft II. Prefers a micro-focused playstyle, often using harassment. \\
Strategic preference & Micro-Focused, Harass Oriented \\
Cognitive heuristics & \textbf{Adaptive Heuristic:} The persona prioritizes harassment if it was successful previously. They adjust their strategy based on observed opponent actions, with a delay. \newline \textbf{Representativeness:} The persona expects the opponent to struggle against harassment, even if they have shown they can defend it. They make assumptions about the opponent's strategy based on limited information. \newline \textbf{Salience:} The persona overreacts to visible enemy units, often neglecting their macro. They are easily distracted by enemy movements. \\
Belief updating & Moderately responsive to new information. Adapts to opponent's strategy but may be slow to recognize when harassment is ineffective. \\
Example response & ``My drops did a lot of damage last game. I'm going to focus on drops again this game and try to keep them on their toes.'' \\
\bottomrule
\end{tabular}
\end{table}

\begin{table}[tbp]
\centering
\caption{Persona Card SC2\_Expert\_1 (expert, macro-oriented profile).}
\label{tab:app-sc2-expert-1}
\small
\begin{tabular}{p{0.21\linewidth}p{0.71\linewidth}}
\toprule
Attribute & Description \\
\midrule
Name & SC2\_Expert\_1 \\
Scenario & \textit{StarCraft~II} \\
Skill level & Expert \\
APM range & 200+ (average: 220) \\
Description & A highly skilled player with extensive knowledge of StarCraft II, capable of executing complex strategies and adapting quickly to the opponent. Prefers a macro-oriented playstyle, aiming for a strong late game. \\
Strategic preference & Macro-Oriented, Late Game Focus \\
Cognitive heuristics & \textbf{Expert Intuition:} The persona quickly determines the optimal macro build order based on the map and matchup. They rapidly assess the game state and make near-optimal decisions. \newline \textbf{Pattern Recognition:} The persona identifies deviations from standard play and adjusts their strategy accordingly. They quickly recognize and exploit weaknesses in the opponent's strategy. \newline \textbf{Planning Fallacy:} While expert at long-term planning, the persona may still underestimate the time needed to make significant tech switches. \\
Belief updating & Rapidly updates beliefs based on new information. Accurately assesses opponent's strategy and adapts accordingly. \\
Example response & ``They're going for a fast expand, so I'll pressure early to punish. I should be able to gain an advantage if I can delay their third base. I need to be careful not to overextend, though.'' \\
\bottomrule
\end{tabular}
\end{table}

\begin{table}[tbp]
\centering
\caption{Persona Card SC2\_Expert\_2 (expert, early-aggression profile).}
\label{tab:app-sc2-expert-2}
\small
\begin{tabular}{p{0.21\linewidth}p{0.71\linewidth}}
\toprule
Attribute & Description \\
\midrule
Name & SC2\_Expert\_2 \\
Scenario & \textit{StarCraft~II} \\
Skill level & Expert \\
APM range & 200+ (average: 250) \\
Description & A highly skilled player with extensive knowledge of StarCraft II, capable of executing complex strategies and adapting quickly to the opponent. Prefers a micro-oriented playstyle with early aggression. \\
Strategic preference & Micro-Oriented, Early Aggression \\
Cognitive heuristics & \textbf{Expert Intuition:} The persona executes precise early game builds. They rapidly assess the game state and make near-optimal decisions. \newline \textbf{Pattern Recognition:} The persona spots holes in the opponent's defenses for harassment. They quickly recognize and exploit weaknesses in the opponent's strategy. \newline \textbf{Confirmation Bias:} The persona interprets early damage as a sign of guaranteed victory, potentially leading to overconfidence. \\
Belief updating & Rapidly updates beliefs based on new information. Accurately assesses opponent's strategy and adapts accordingly. \\
Example response & ``Their build is greedy, I can punish them with lings. I need to make sure I don't lose too many units, though, or I'll be behind.'' \\
\bottomrule
\end{tabular}
\end{table}

We embedded researcher judgments in the cards, including assumed APM bands, cognitive heuristics, and normative interpretations of strategy. We treat these fields as contestable hypotheses rather than measured player attributes. We included no demographic fields in the six transcribed cards; because we did not retain complete prompts and persona backstories, we do not claim that the broader generation process was demographically neutral.

\appendix
\clearpage
\endgroup

\end{document}